%% file: paper.tex
\documentclass[sigconf]{acmart}
\usepackage{booktabs} 

\input{macro}

\copyrightyear{2023}
\acmYear{2023}
\setcopyright{acmlicensed}\acmConference[SA Technical Communications '23]{SIGGRAPH Asia 2023 Technical Communications}{December 12--15, 2023}{Sydney, NSW, Australia}
\acmBooktitle{SIGGRAPH Asia 2023 Technical Communications (SA Technical Communications '23), December 12--15, 2023, Sydney, NSW, Australia}
\acmPrice{15.00}
\acmDOI{10.1145/3610543.3626162}
\acmISBN{979-8-4007-0314-0/23/12}

\begin{document}

\def\datasetname{MicroGlam}

\title[MicroGlam: Microscopic Skin Image Dataset with Cosmetics]{MicroGlam: Microscopic Skin Image Dataset with Cosmetics}

\author{Toby Chong}
\orcid{0000-0001-5992-5647}
\email{tobyclh@gmail.com}
\affiliation{%
 \institution{The University of Tokyo}
 \country{Japan}}

\author{Alina Chadwick}
\orcid{0009-0006-3016-7663}
\email{alina.chadwick@gmail.com}
\affiliation{%
 \institution{Dartmouth College}
 \country{USA}}

\author{I-Chao Shen}
 \orcid{0000-0003-4201-3793}
 \email{ichaoshen@g.ecc.u-tokyo.ac.jp}
 \affiliation{%
  \institution{The University of Tokyo}
  \country{Japan}}
 
\author{Haoran Xie}
  \orcid{1234-5678-9012-3456}
  \email{xie@jaist.ac.jp}
  \affiliation{%
   \institution{JAIST}
   \country{Japan}}
 
\author{Takeo Igarashi}
\orcid{0000-0002-5495-6441}
\email{takeo@acm.org}
\affiliation{%
\institution{The University of Tokyo}
\country{Japan}}
   
\input{abstract}



\begin{teaserfigure}
  \includegraphics[width=\textwidth]{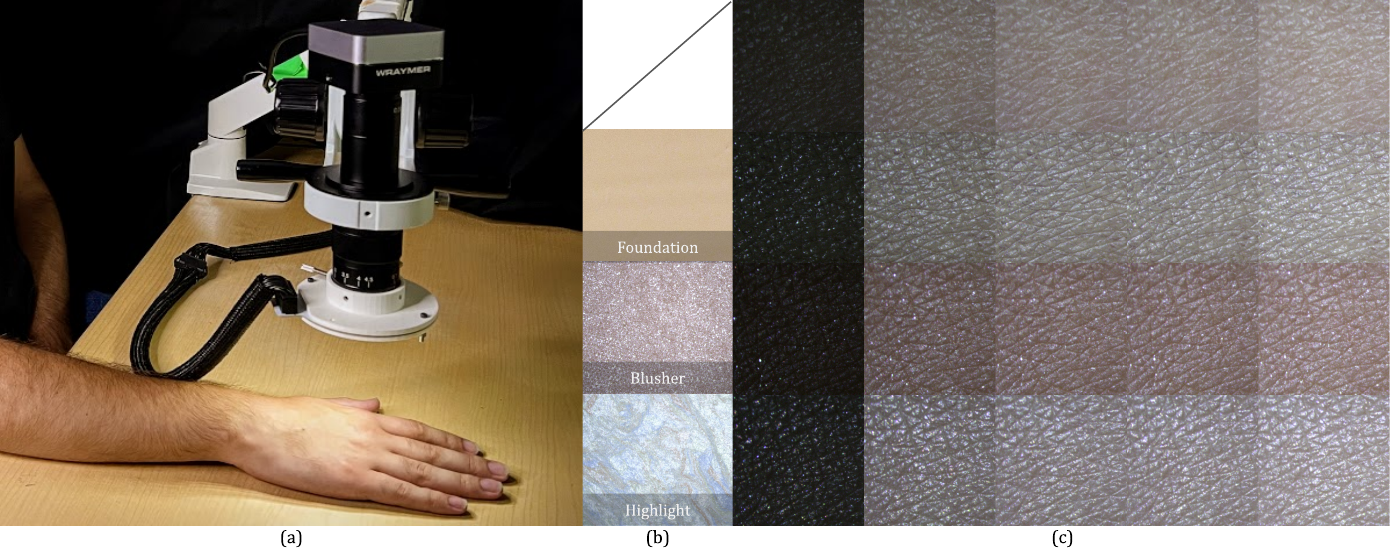}
  \caption{
    We developed (a) a capturing device consisting of a microscopic lens and $16$ LEDs to capture a skin patch image dataset with cosmetics on the back of hand region.
    This dataset consists of $8mm^*8mm$ skin patch images under three cosmetic products (foundation, blusher, and highlighter).
    (c) We showed images captured under diverse lighting conditions for each cosmetic product.
  }
  \label{fig:teaser}
\end{teaserfigure}

\maketitle

\input{introduction}
\input{relatedwork}

\input{method}
\input{application}

\input{experiment}
\input{conclusion}

\bibliographystyle{ACM-Reference-Format}
\bibliography{paper}

\end{document}

%% file: macro.tex
\usepackage{enumitem}
\usepackage{amsmath}
\usepackage{amsfonts}
\usepackage{multirow}
\usepackage{bigdelim}
\usepackage{color}
\usepackage{booktabs}
\usepackage{ifthen}
\usepackage{hyperref}
\usepackage{subcaption}
\usepackage{bbold}
\usepackage{xcolor, pgf}
\usepackage{color, colortbl}
\usepackage[normalem]{ulem} 
\usepackage{placeins}
\usepackage{soul}
\usepackage[skins]{tcolorbox}
\usepackage[noabbrev,capitalise,nameinlink]{cleveref}
\creflabelformat{equation}{#2\textup{#1}#3}

\usepackage{algorithm}
\usepackage{algorithmicx}
\usepackage{algpseudocode}
\algnewcommand\algorithmicinput{\textbf{Input:}}
\algnewcommand\INPUT{\item[\algorithmicinput]}
\algnewcommand\algorithmicoutput{\textbf{Output:}}
\algnewcommand\OUTPUT{\item[\algorithmicoutput]}
\algnewcommand\algorithmicforeach{\textbf{for each}}
\algtext*{EndFor}

\algdef{S}[FOR]{ForEach}[1]{\algorithmicforeach\ #1\ \algorithmicdo}
\algrenewcommand{\alglinenumber}[1]{\color{red!80!blue}\footnotesize#1:}

\algnewcommand\Func[2]{\textcolor{green}{#1}\textcolor{green}{(#2)}}
\algnewcommand\Insert[2]{Insert {#1} to #2.}
\algnewcommand\Input[1]{\State \textbf{Input: } #1}
\algnewcommand\Output[1]{\State \textbf{Output: } #1}

\newlength\myboxwidth
\definecolor{gray}{rgb}{0.5,0.5,0.5}
\definecolor{green}{rgb}{0, 0.6, 0}
\definecolor{orange}{rgb}{1, 0.5, 0}
\definecolor{mahogany}{rgb}{0.75, 0.25, 0.0}
\definecolor{purple}{rgb}{0.6, 0, 0.6}
\definecolor{darkgreen}{rgb}{0, 0.3, 0}
\definecolor{orange}{rgb}{1, 0.5, 0.}
\definecolor{lightblue}{rgb}{0.52, 0.75,0.91}
\definecolor{softgreen}{rgb}{0.66,0.87,0.74}
\definecolor{softred}{rgb}{0.96,0.71,0.69}

\newcommand{\ignore}[1]{}
\newcommand{\none}[1]{}
\newcommand{\com}[1]{}

\newcommand{\etal}{{\it{et~al.}}}
\newcommand{\ie}{i.e.,}
\newcommand{\eg}{e.g.,}

%% file: abstract.tex
\begin{abstract}
In this paper, we present a cosmetic-specific skin image dataset.\footnote{https://github.com/tobyclh/MicroGlam} 
It consists of skin images from $45$ patches ($5$ skin patches each from $9$ participants) of size $8mm^*8mm$ under three cosmetic products (\ie~foundation, blusher, and highlighter). 
We designed a novel capturing device inspired by LightStage~\cite{debevec2000acquiring}. 
Using the device, we captured over $600$ images of each skin patch under diverse lighting conditions in $30$ seconds. 
We repeated the process for the same skin patch under three cosmetic products.
Finally, we demonstrate the viability of the dataset with an image-to-image translation-based pipeline for cosmetic rendering and compared our data-driven approach to an existing cosmetic rendering method~\cite{kim2018practical}.
\end{abstract}

%% file: introduction.tex
\section{Introduction}
\label{sec:intro}


We examine existing cosmetic image datasets~\cite{li2018beautygan, gu2019ladn, dantcheva2012youtube_ymu}, which are often built to enable copying a general ``look'' from a face portrait with makeup to another without. 
They often consist of face portraits with and without makeup, mostly from different individuals.
Utilizing the aforementioned datasets fundamentally entangles multiple challenging problems, such as separating the cosmetics from skin tone and lighting as well as separating the compounded effect of different cosmetic products when they are applied on top of each other or in multiple layers. 
The lack of specificity of the particular cosmetic product being applied onto portraits in most cosmetic datasets also limits the use of them to answer a very fundamental question, \textit{``How would I look in this specific product?''}

We aim to simplify the problem to what we call \textbf{cosmetic rendering}, where a known cosmetic product is virtually applied onto an image of skin. 
We do so by creating the first cosmetic-specific skin image dataset \textit{\textbf{\datasetname}}, a dataset of microscopic skin patches with and without cosmetics, designed and collected with the following goals in mind:
\begin{itemize}
    \item contains the makeup and no-makeup data to accurately isolate the effect of a cosmetic of interest on skin, 
    \item allows further processes to extract not only the diffuse color but also other visual parameters from the cosmetics, 
    \item contains important metadata, including the cosmetic used, the method of application, and the skin tone of the subject. 
    
\end{itemize}

We first developed a capturing and data processing framework, which contains $16$ LEDs attached onto a microscope that lights the skin patch from $16$ directions for accurate measurement. With this device, we captured skin patches at the back of the hand region, where cosmetic products (including foundation, blusher, and highlighter) are often tested in store. 
Using the proposed device, we captured data from $9$ subjects aged $20$-$32$. 
For each participant, we captured $600$ images for each of five distinct skin patches at the back of the hand region.
Each individual skin patch was captured under three different cosmetic products along with a reference capture in its natural, no-makeup state (\cref{table:dataset}). 
We demonstrated that our dataset can be leveraged to accurately render cosmetics onto an unseen skin patch with an unknown skin-tone and are of greater realism in comparison to the results generated by an existing cosmetic transfer work~\cite{kim2018practical}. 

%% file: relatedwork.tex
\section{Related Work}
\label{sec:related}

\subsection{Cosmetic datasets}
Cosmetic datasets are often curated in the form of ``unpaired'' datasets~\cite{li2018beautygan, gu2019ladn}, where makeup and no-makeup portraits belong to different individuals. 
On the other hand, ``paired'' datasets, which include makeup and no-makeup portraits of the same person are rare.
Dantcheva~\etal~\shortcite{dantcheva2012youtube_ymu} created a paired dataset from YouTube makeup tutorials. The dataset is paired as it contains before and after makeup images of the same person, yet the faces are often under different lighting conditions and poses and the resolution of the videos cannot capture the subtle changes of individual products. Synthetic data is another popular approach to generate paired data by applying virtual makeup to no-makeup portraits~\cite{sajid2018data} and algorithmic makeup removal on makeup portraits~\cite{yang2023makeup_cyber}.
Scherbaum~\etal~\shortcite{scherbaum2011computer} captured paired full face makeup data using a multi-light device similar to Lightstage, yet they only captured data from three people with no documentation regarding the cosmetic product(s) applied.



\subsection{Detailed skin appearance capturing}
Reproducing realistic skin appearance (both texture and geometry) is key to creating realistic virtual makeup. 
However, it remains challenging and requires specialized setups due to the complex physical and visual properties of skin, \eg non-rigid shape, microgeometry (wrinkles) and, subsurface scattering. 
One way to capture the microgeometry is to cast a mold onto the skin area~\cite{haro2001real}. 
However, this method cannot capture the interaction between the skin and cosmetics. 


%% file: method.tex
\section{Capturing procedure and data processing}\label{sec:method}
In this section, we discuss the design of our capturing device, the capturing procedure, and the data processing pipeline.

\begin{figure}
    \centering
    \includegraphics[width=\columnwidth]{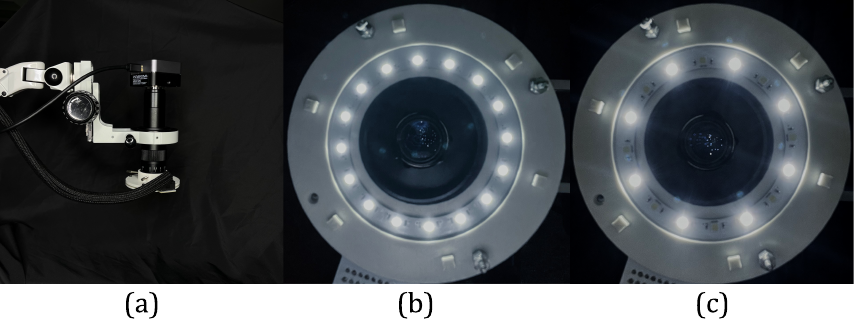}{}
    \caption{(a) Side view of our capturing device. (b)(c) Bottom view of the device with all and half of the LEDs selectively turned on, respectively. Each of the $16$ LEDs surrounding the camera center can be switched on/off individually.}
    \label{fig:device}
\end{figure}

\begin{figure*}
    \centering
    \includegraphics[width=\textwidth]{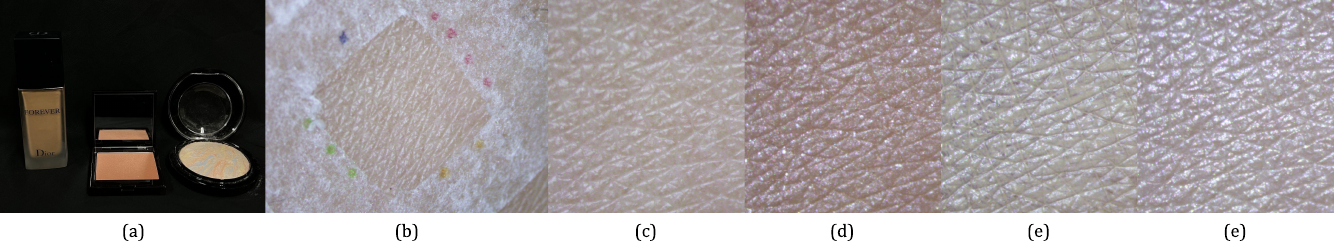}
    \caption{(a) Cosmetic products captured: foundation, blusher, and highlighter. (b) Raw image of a skin patch captured using our device of approximately $8mm^*8mm$. (c) Cropped and aligned version of (b). (d, e, f) The same skin patch but with blusher, foundation, and highlighter applied onto it respectively. We recommend enlarging the figures for a clearer viewing experience. Note how all makeup interacts with the skin microgeometry and affects more than just the diffuse color of the skin (\eg~foundation fills in the wrinkles of the skin and highlighter adds significant specular reflection onto the surface of skin).}
    \label{fig:makeup_sample}
\end{figure*}

\subsection{Capturing device}
We had the first observation found that cosmetic products heavily interact with skin microgeometry (\ie~fine wrinkles and pores). We hence determined that microscopic data is required if we aimed to reproduce subtle changes to the skin. 
To this end, we based our capturing device on a digital microscope (Wraycam VEX120) with a native resolution of $1280^*960$. We capture skin at roughly $7,000$ pixel/cm (for scale, a 4K face portrait is roughly $200$ pixel/cm). 
A 3D-printed ring light with $16$ LEDs is placed uniformly surrounding the microscope center with a radius of $5mm$. 
The state of the lights (on/off) can be controlled individually to simulate complex lighting conditions from one or multiple directions. 
We controlled both the camera and the ring light at $20$FPS to ensure the lights and camera at synchronized during our capturing procedure. 

\subsection{Subjects and cosmetic products}
We recruited $9$ subjects aged from $20$-$32$.
Their skin tones range from shade three to six in the MST scale~\cite{monk2023monk}. 
We sampled five patches from each of the subjects' hands. 
We captured their skin with highlighter (M.A.C. Mineralize Skinfinish Lightscapade), blusher (Laura Mercier blush color infusion 05 Shimmer), and foundation (Christian Dior Forever skin glow 2N) as well as a reference no-makeup capture.

\begin{table}[h!]
\begin{tabular}{|l|l|}
\hline
\ (a) Subjects & 9  \\ \hline
\ (b) Cosmetic products   & \begin{tabular}[c]{@{}l@{}}No-makeup (reference), foundation, \\ blusher, highlighter\end{tabular} \\ \hline
\ (c) Patches per person & five $8mm^*8mm$ skin patches \\ \hline
\ (d) Images per sequence & \begin{tabular}[c]{@{}l@{}}600 images for randomized lights and \\ 16 images for one-light at a time \end{tabular} \\ \hline
\textbf{Total image count} & $(a)^*(b)^*(c)^*(d)$ = 110880 \\ \hline
\end{tabular}
\caption{Dataset overview. }
\label{table:dataset}
\end{table}

\subsection{Capturing Procedure}\label{sec:capture_procedure}
At the beginning of the capture session, we prepared five registration markers per subject that consisted of a $18mm^*18 mm$ patch of surgical tape with a $8mm^*8 mm$ hole in each patch to expose the subject's skin. On the borders of the hole, we added coloured dots using marker pens for image alignment~(\cref{fig:makeup_sample}b).
We applied the registration markers to one hand and captured a \textit{sequence} of $616$ images of the skin patch under each marker without cosmetic product, and repeat the process for the three cosmetic products. 
When applying highlighter or blush, we used designated brushes dipped in a small amount of product (enough to cover the patch) to apply the makeup. When applying foundation, we squeezed out product onto a palette and used a designated brush to apply small amounts of foundation to the patch of interest.

Within a sequence, we randomly varied both the intensity of the lights as well as the light combinations, with $3$ to $8$ of the $16$ lights on at a time for $600$ images plus additional $16$ images with one light on at a time for a total of $616$ images.  
Notice that the lighting conditions of all captured sequences are identical across patches and makeup, and only varies among different subjects. 
Each capture sequence took approximately $30$ seconds, and the entire capturing session took around $45$ minutes for five patches and four rounds of capturing.

\subsection{Skin patch alignment}


After capturing, we manually aligned all images of the same patch of skin with makeup to those without makeup. 
We manually identify correspondences in the first images of two sequences and compute the homography transformation. We applied the same transformation to all images from the makeup sequences to align to the no-makeup sequence. 


%% file: application.tex
\begin{figure*}[h!]
    \centering
    \includegraphics[width=\textwidth]{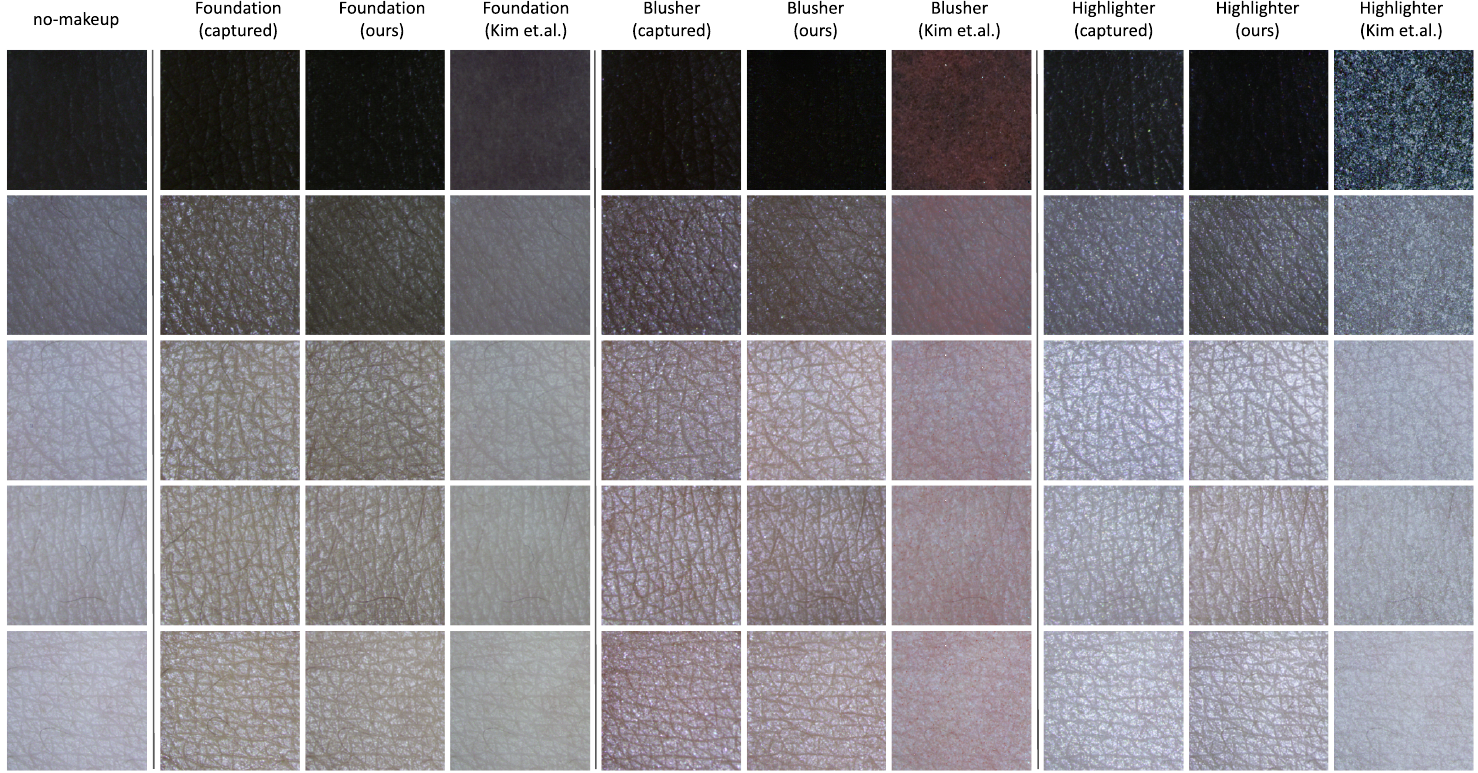}{}
    \caption{Cosmetic rendering results generated by our image translation-based approach (\cref{sec:app}) and the baseline method (\cite{kim2018practical}). 
    We used the no-makeup images captured at different lighting conditions as input to generate all the results. 
    For each cosmetic product, we provide a reference by presenting the captured skin patch with the exact same cosmetic product applied onto it.
    }
    \label{fig:app}
\end{figure*}
\section{Application - Cosmetic Rendering}
\label{sec:app}
To demonstrate the potential of our dataset, we propose a cosmetic rendering application based on the image-to-image translation network~\cite{zhu2017unpaired_cyclegan}.
The goal is to estimate the appearance of a given skin patch when a specific cosmetic product is applied to it.
Using the data we captured in~\cref{sec:method}, we constructed three sub-datasets for each target cosmetic products: \textit{foundation}, \textit{blusher}, and \textit{highlighter}. 
Each sub-datasets consisted of paired skin patches captured both with and without makeup application.
We merged captured skin patches from all subjects to improve the skin tone variation of each sub-datasets.
For each target cosmetic product, we trained a separate network using the corresponding sub-dataset.
During inference, given the appearance of the skin without any cosmetics, each trained model predicts the appearance of skin patches under the application of the corresponding cosmetic.

\begin{figure}
    \centering
    \includegraphics[width=\columnwidth]{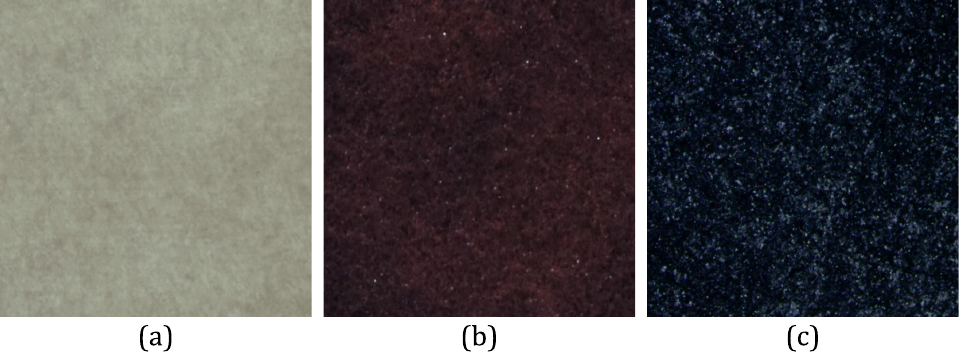}{}
    \caption{Following Kim~\etal, we applied (a) foundation onto a white surface and (b) blusher and (c) highlighter onto a black surface. We use them as input to the baseline method. 
    (zoomed-in images of different cosmetic products)
    }
    \label{fig:baseline_app}
\end{figure}

In addition, we compare our cosmetic rendering result with that generated using the method proposed by Kim~\etal~\shortcite{kim2018practical} (\textit{baseline}). 
The baseline method transfers cosmetic product applied onto a flat surface (\cref{fig:baseline_app}) to a skin patch. 
We chose the thickness parameter in the baseline method such that it minimizes the error between the baseline result and the ground truth.   

In~\cref{fig:app}, we demonstrate all five patches captured from a subject whose data is not used during training, under varied lighting conditions. 
We showed that using our dataset, our method can generate convincing cosmetic rendering results that are closer to the captured skin patches with cosmetics applied to them.
In particular, our method preserves the microgeometry of the input skin patch, while generating the desired specular effects after applying different cosmetic products.

%% file: experiment.tex


%% file: conclusion.tex
\section{Conclusion}
In this work, we presented \datasetname, a dataset of microscopic skin patches with and without cosmetic product.
This dataset is designed and collected to facilitate the development of cosmetic rendering. 
It contains a variety of patches under cosmetic products.
We demonstrated the potential of using \datasetname~to perform accurate cosmetic rendering. 
As future directions, we would expand on the diversity of our dataset in terms of cosmetics and skin-tones of subjects, and include skin data captured on the face. 